\documentclass[final]{cvpr}

\usepackage{times}
\usepackage{epsfig}
\usepackage{graphicx}
\usepackage{amsmath}
\usepackage{amssymb}

\usepackage{multirow}
\usepackage{blindtext}
\usepackage{placeins}
\usepackage{adjustbox}
\usepackage{xfrac}
\usepackage{enumitem}

\usepackage{array}
\usepackage{rotating}
\usepackage{multicol}
\usepackage{subfigure}
\usepackage{chngcntr}
\usepackage{bm}
\usepackage{booktabs}

\usepackage[pagebackref=true,breaklinks=true,colorlinks,bookmarks=false]{hyperref}

\def\confYear{CVPR 2021}
\begin{document}

\title{Selective Replay Enhances Learning in Online Continual Analogical Reasoning}

\author{Tyler L. Hayes$^1$ and Christopher Kanan$^{1,2,3}$\\
Rochester Institute of Technology$^1$, Paige$^2$, Cornell Tech$^3$\\
{\tt\small \{tlh6792, kanan\}@rit.edu}\\
}

\maketitle

\begin{abstract}
    In continual learning, a system learns from non-stationary data streams or batches without catastrophic forgetting. While this problem has been heavily studied in supervised image classification and reinforcement learning, continual learning in neural networks designed for abstract reasoning has not yet been studied. Here, we study continual learning of analogical reasoning. Analogical reasoning tests such as Raven's Progressive Matrices (RPMs) are commonly used to measure non-verbal abstract reasoning in humans, and recently offline neural networks for the RPM problem have been proposed. In this paper, we establish experimental baselines, protocols, and forward and backward transfer metrics to evaluate continual learners on RPMs. We employ experience replay to mitigate catastrophic forgetting. Prior work using replay for image classification tasks has found that selectively choosing the samples to replay offers little, if any, benefit over random selection. In contrast, we find that selective replay can significantly outperform random selection for the RPM task\footnote{\url{https://github.com/tyler-hayes/Continual-Analogical-Reasoning}}. 
\end{abstract}

\section{Introduction}
\label{sec:intro}

\begin{figure}[t]
\begin{center}
  \includegraphics[width=0.75\linewidth]{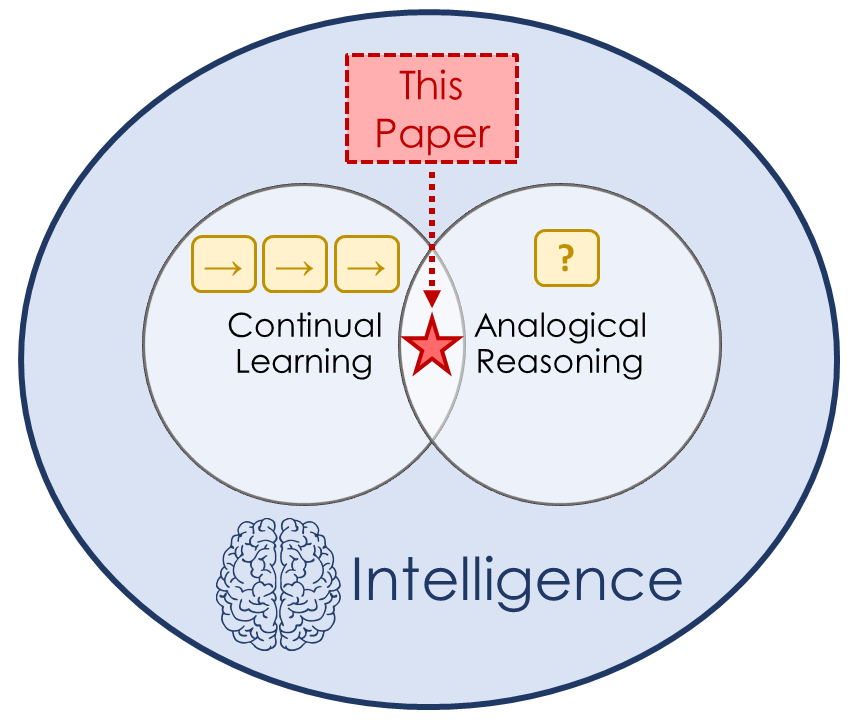}
\end{center}
  \caption{Humans are capable of continual learning, analogical reasoning, and performing both simultaneously. This facilitates much of what we consider intelligence. However, conventional neural networks suffer from catastrophic forgetting when updated on non-stationary data distributions and struggle to reason abstractly. Here, we study a network's ability to continually learn analogical reasoning tasks.
  }
\label{fig:main}
\end{figure}

Deep neural networks excel at pattern recognition tasks, and they now rival or surpass humans at tasks such as image classification. However, the human capability for image classification is not unique in the animal kingdom. Multiple primate species are also capable of image classification at levels that rival humans~\cite{fagot2006evidence,rajalingham2018large,rajalingham2015comparison}. One of the characteristics of human intelligence that distinguishes us over other animals is our ability to perform analogical reasoning~\cite{gentner2001analogical,gentner1997structure,lovett2017modeling}. Specifically, analogical encoding, i.e., the comparison of two situations which are partially understood, has been shown to facilitate forward knowledge transfer to new problems, as well as backwards transfer for memory retrieval~\cite{gentner2004analogical}. This transfer is facilitated through the formation of higher level abstract schemas over time that are derived from comparisons between situations. More generally, abstraction capabilities enable forward knowledge transfer in humans, where previously learned knowledge is used to improve future learning~\cite{gentner2009reviving,ho2019value}. Recently, multiple deep learning models for analogical reasoning have been proposed~\cite{barrett2018measuring,spratley2020closer,teney2020v,zheng2019abstract}; however, these systems assume that all training data is available at once and they will not be updated again, so backward and forward transfer in these systems cannot be studied. In this paper, we pioneer the development of the first systems for online continual learning of analogical reasoning tasks over time enabling forward and backward transfer in these tasks to be studied in models (see Fig.~\ref{fig:main}).

\begin{figure}[t]
\begin{center}
  \includegraphics[width=0.85\linewidth]{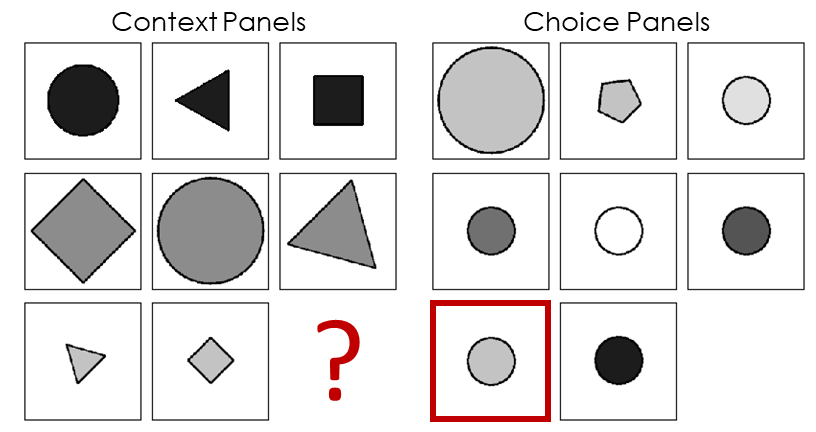}
\end{center}
  \caption{Example of an RPM problem from RAVEN~\cite{zhang2019raven}. Each RPM problem consists of 8 context panels (images) and 8 choice panels (images). Given these 16 panels as inputs, an agent must reason about the relationships among the 8 context panels and choose the best choice panel to complete the context matrix. For this example, we highlighted the answer panel with a red box.
  }
\label{fig:rpm-example}
\end{figure}

While humans have strong abstract reasoning abilities, neural networks struggle with these problems~\cite{hernandez2016computer,hill2018learning,barrett2018measuring,saxton2018analysing,zhang2019raven}. A popular cognitive test for analogical reasoning in humans is the Raven's Progressive Matrices (RPMs) problem~\cite{ea1938raven}, which provides a user with a 3$\times$3 grid of images where the last image in the grid is missing (see Fig.~\ref{fig:rpm-example}). The rows and/or columns of images in the grid follow a specific rule and the user must compare a set of 8 choice images to select the one that best fits in the final location. RPMs are ideal for measuring analogical reasoning in neural networks because they isolate the reasoning task directly, and several RPM datasets for neural networks have been created~\cite{barrett2018measuring,zhang2019raven}.

Network architectures have been proposed for the RPM problem, with the most successful using a form of relation network~\cite{barrett2018measuring,spratley2020closer,teney2020v,zheng2019abstract}. These models perform well in an offline setting, where they are trained on all available data and then evaluated. However, offline learning means that forward and backward transfer in these systems cannot be studied, and this is one of the critical capabilities that abstract reasoning is thought to facilitate in humans. When conventional neural networks are trained incrementally from non-stationary data streams, mirroring human learning, they suffer from catastrophic forgetting~\cite{mccloskey1989}. The field of continual learning endeavors to overcome this limitation~\cite{belouadah2020comprehensive,de2019continual,kemker2018forgetting,parisi2019continual}. 
\textbf{This paper makes the following contributions:}
\begin{itemize}[noitemsep, nolistsep]
    \item We pioneer continual learning for analogical reasoning problems. We establish protocols and metrics for this problem using the Relational and Analogical Visual rEasoNing (RAVEN) dataset~\cite{zhang2019raven}.
    
    \item We integrate both regularization and replay continual learning mechanisms into neural networks for analogical reasoning to establish baseline results. 
    
    \item We study replay selection policies and find improved performance over uniform random selection for the RPM problem. This is interesting as selective replay has shown little benefit over uniform random selection in standard supervised classification settings~\cite{chaudhry2018riemannian,hayes2019remind}.

\end{itemize}

\section{Problem Formulation}
\label{sec:problem-formulation}

There are two paradigms for training a continual learner: the incremental batch paradigm and the online streaming paradigm. In incremental batch learning of RPMs, a dataset is divided into $T$ batches, i.e., $\mathcal{D}=\bigcup_{t=1}^{T}B_{t}$, and at each time-step, $t$, the learner receives a new batch of data consisting of $N_t$ samples, i.e., $B_{t}=\left\{\left(\mathbf{S}_{i},y_{i}\right)\right\}_{i=1}^{N_{t}}$, where $\mathbf{S}_{i}$ is a single RPM sample consisting of 8 context panels (images), 8 choice panels (images), and a label $y_{i}$ indicating which choice panel is the correct answer. Given this batch of samples, the agent is allowed to loop over the batch until it has been learned and is subsequently evaluated. The sequences of batches are ordered by task, which would induce catastrophic forgetting in a conventional network.

\begin{figure*}[t]
    \centering
        \includegraphics[width=0.75\textwidth]{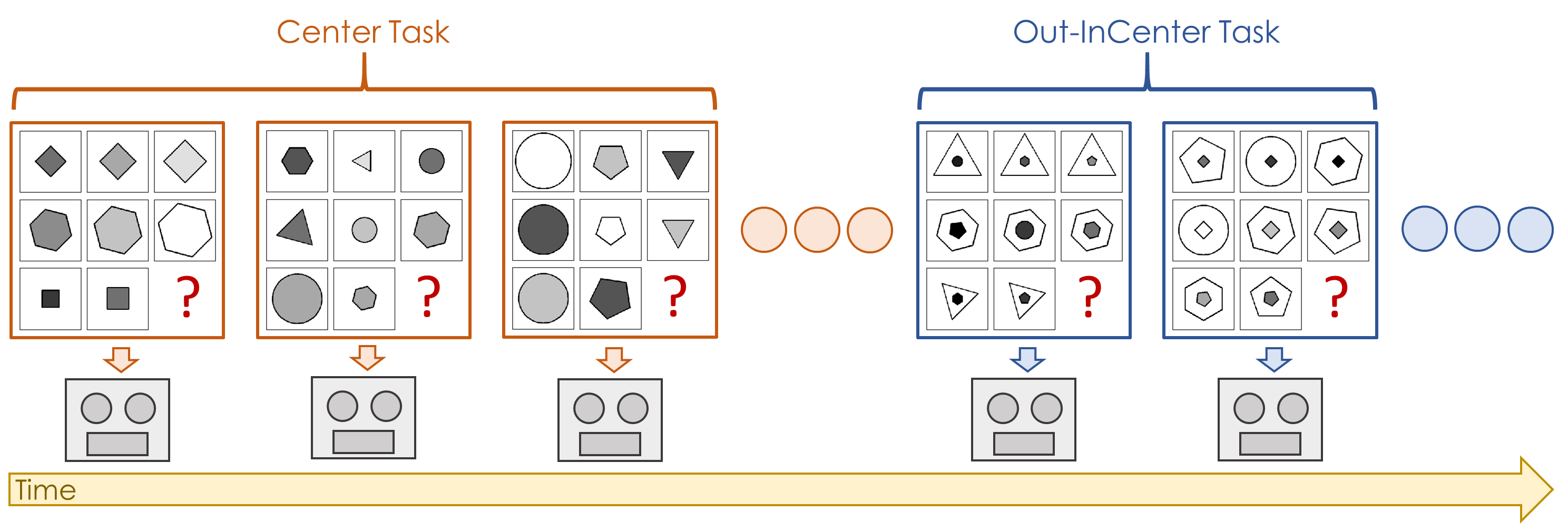}
    \caption{A demonstration of our training protocol on RAVEN~\cite{zhang2019raven}. At each time-step, a streaming learner receives a single new example from a particular task. It learns the new example and then can be subsequently evaluated.
    }
    \label{fig:rpm-protocol}
\end{figure*}

The incremental batch paradigm is unrealistic for real-time agents that must learn and evaluate on new data immediately and it does not mirror human learning. Online streaming learning addresses these drawbacks and requires an agent to learn new samples one at a time ($N_{t}=1$) with only a single epoch through the entire dataset. Further, it requires models to operate under severe memory and time constraints, making them more ideal for deployment. Our streaming protocol is depicted in Fig.~\ref{fig:rpm-protocol}.

\section{Related Work}
\label{sec:related-work}

\subsection{Neural Networks for RPMs}
\label{subsec:related-analogical}

One of the earliest works on neural networks for RPMs~\cite{carpenter1990one} identified a discrete set of rules required to solve them. This set of rules was used in \cite{wang2015automatic} to develop a technique for automatically generating valid RPM problems, which was used to create the Procedurally Generated Matrices (PGM) dataset~\cite{barrett2018measuring}. PGM was the first large-scale RPM-based dataset containing enough problems to successfully train deep neural networks. The Wild Relation Network (WReN) for PGM was proposed in \cite{barrett2018measuring}, which uses a relation network~\cite{santoro2017simple} for reasoning. WReN outperformed other baselines on PGM by over 14\%, demonstrating the strength of relation networks for analogical reasoning. Performance has been further improved by augmenting WReN with an unsupervised disentangled variational autoencoder~\cite{steenbrugge2018improving} and Transformer attention mechanisms~\cite{hahne2019attention}. 

More recently, the synthetically generated RAVEN dataset was released~\cite{zhang2019raven}. Unlike PGM, each image in RAVEN contains objects in a structured pattern, requiring models to perform both structural and analogical reasoning. While WReN works well on PGM, it performs much worse on RAVEN, which is attributed to its lack of compositional reasoning abilities. To improve performance on RAVEN, the Contrastive Perceptual Inference Network uses a contrastive loss and contrast module to jointly learn rich features for perception~\cite{zhang2019learning}. In \cite{zheng2019abstract}, a reinforcement learning policy was used to select informative samples for training a Logic Embedding Network to reason about panels. In \cite{Wang2020Abstract}, a graph neural network was used to extract object representations from images and reason about them.

Today, the Rel-Base and Rel-AIR architectures have the best results on RAVEN and they perform competitively on PGM~\cite{spratley2020closer}. Rel-Base uses an object encoder network to process image panels individually. These encodings are passed to a sequence encoder to extract relationships before being scored. Rel-AIR first trains the Attend-Infer-Repeat (AIR) module~\cite{eslami2016attend} to extract objects from images. These objects are encoded and paired with additional position and scale information before being processed by a sequence encoder. Due to Rel-Base's simplicity and strong results, we extend it to the continual learning setting here. 

\subsection{Continual Learning in Neural Networks}
\label{subsec:related-continual}

One of the hallmarks of human intelligence is our ability to continually learn new information throughout our lifetimes without catastrophically forgetting previous knowledge. However, neural networks struggle to learn from non-stationary data distributions over time. When naively updated on new information, conventional neural networks catastrophically forget prior information~\cite{french1999catastrophic,mccloskey1989}, which results from the stability-plasticity dilemma~\cite{abraham2005memory,mccloskey1989}. The field of continual learning seeks to overcome these challenges.

There are three primary mechanisms used to mitigate forgetting in neural networks~\cite{belouadah2020comprehensive,de2019continual,kemker2018forgetting,parisi2019continual}: 1) regularizing the plasticity of weights~\cite{aljundi2018memory,chaudhry2018riemannian,chaudhry2019efficient,dhar2019learning,kirkpatrick2017,li2016learning,lopez2017gradient,ritter2018online,serra2018overcoming,zenke2017continual}, 2) growing the network architecture to accommodate new data~\cite{hayes2019lifelong,hou2018lifelong,ostapenko2019learning,rusu2016progressive,yoon2018lifelong}, and 3) maintaining a subset of previous data in a memory buffer or generating previous data to replay when new data becomes available~\cite{belouadah2019il2m,castro2018end,chaudhry2019efficient,douillard2020podnet,french1997pseudo,hayes2019memory,hayes2019remind,hou2019unified,kemker2018fearnet,rebuffi2016icarl,tao2020topology,wu2019large}. Specifically, several regularization strategies seek to directly preserve important network parameters over time~\cite{aljundi2018memory,chaudhry2018riemannian,kirkpatrick2017,zenke2017continual}, while others use variants of distillation to preserve outputs at various locations in the network~\cite{dhar2019learning,douillard2020podnet,li2016learning}. However, for image classification, replay (or rehearsal) methods are currently the state-of-the-art approach, especially for large-scale datasets~\cite{castro2018end,douillard2020podnet,hayes2019remind,hou2019unified,rebuffi2016icarl,tao2020topology,wu2019large}. Standard replay approaches store explicit images or features in a memory buffer, while generative replay (or pseudo-rehearsal) methods train a generative model to generate previous samples~\cite{french1997pseudo,he2018exemplar,kemker2018fearnet,ostapenko2019learning,shin2017continual}. When new data becomes available, either all or a subset of these previous samples are mixed with new samples to fine-tune the network.

Continual learning has also been studied for object detection~\cite{acharya2020rodeo,shmelkov2017incremental}, semantic segmentation~\cite{cermelli2020modeling,michieli2019incremental}, and robotics~\cite{feng2019challenges,lesort2020continual}. Especially relevant to this paper are continual learners for visual question answering~\cite{greco-2019-psycholinguistics,hayes2019remind}, which requires agents to answer questions about images. However, visual question answering requires models to process natural language inputs, overcome severe dataset biases, and does not provide an isolated test of analogical reasoning. Using RPMs enables us to isolate continual analogical reasoning before moving on to tasks requiring more abilities.

\section{Continual Learning Models \& Baselines}
\label{subsec:baselines}

We train continual models on RPM tasks in the following way. First, we perform a base initialization phase where Rel-Base is trained offline on the first task. After base initialization, each continual learner starts from these initialization weights and learns each of the remaining tasks one at a time, while being evaluated on all test data after every task. While batch models process new tasks in batches that they loop over several times, streaming models process samples one at a time, with only a single epoch through the dataset. For all models, we fix the task and sample order.

We study three methods to enable continual learning in Rel-Base: Distillation, EWC, and Partial Replay. We also study three baselines. These are described below:
\begin{itemize}[noitemsep, nolistsep]

    \item \textbf{Fine-Tune} -- Rel-Base without any mechanisms to enable continual learning. It serves as a lower bound on performance. We run fine-tune in the streaming and batch paradigms.
    
    \item \textbf{Distillation} --  Given a new batch of data, Distillation~\cite{hinton2015distilling} optimizes both a classification and distillation loss, where the soft targets of the distillation loss are the scores of the model from the previous time-step. Distillation has been effective in mitigating forgetting on image classification~\cite{dhar2019learning,douillard2020podnet,li2016learning,rebuffi2016icarl}, object detection~\cite{shmelkov2017incremental}, and semantic segmentation~\cite{cermelli2020modeling} tasks.
    
    \item \textbf{EWC} -- The Elastic Weight Consolidation (EWC) batch model uses a quadratic regularization term to encourage weights to remain close to their previous values~\cite{kirkpatrick2017}. Given a new batch of data, EWC optimizes both a classification loss and a quadratic penalty loss weighted by each parameter's importance, determined by the Fisher Information Matrix.
    
    \item \textbf{Partial Replay} -- Partial Replay continually fine-tunes a model with all new data and a \emph{subset} of previous data. It achieves strong results on classification tasks~\cite{castro2018end,hou2019unified,rebuffi2016icarl,wu2019large}. We use it in the streaming setting and provide more details in Sec.~\ref{sec:sample-selection}.
    
    \item \textbf{Cumulative Replay} -- Cumulative Replay continually fine-tunes a model with \emph{all} new and previous data. It has been shown to mitigate forgetting~\cite{hayes2019memory}, but is resource intensive. We train Cumulative Replay in the batch setting due to compute constraints.
    
    \item \textbf{Offline} -- This is an offline model trained from scratch on all data until the current time-step. It serves as an upper bound on continual learning performance.
\end{itemize}

Distillation and EWC operate on batches since their loss constraints are computed from a model at the previous time-step, which would be less beneficial in the streaming setting. For Partial Replay, we study multiple policies for choosing which samples to replay, which we describe next.

\subsection{Selective Replay Policies}
\label{sec:sample-selection}

Studying selective replay is important because it has the potential to enable better network generalization, allow networks to use fewer computational resources, and more closely aligns with biology~\cite{lee_memory_2002,louie_temporally_2001,peyrache_replay_2009}. Although replay selection policies have been explored for supervised classification~\cite{aljundi2019online,aljundi2019gradient,chaudhry2018riemannian,mcclelland2020integration,wu2019large}, they have not yielded significant improvements, especially on large-scale datasets~\cite{hayes2019remind,wu2019large}. For example, \cite{wu2019large} found that selective replay performed 0.46\% better on average over random sampling on CIFAR-100, but it is unclear if this improvement is statistically significant. In reinforcement learning, selective replay has provided more benefit~\cite{andrychowicz2017hindsight,Schaul2016PrioritizedER}. Further, selectively choosing training samples has yielded improved performance for offline RPM models~\cite{hill2018learning,zheng2019abstract}, but its effectiveness in the continual setting has not been explored, so we study it here.

Formally, we train a Partial Replay model in two stages: a base initialization phase and a streaming phase. During base initialization, we train the network offline using standard mini-batches and optimization updates. We then subsequently store all base initialization data in a replay buffer, $\mathcal{B}$. Then, labeled RPM sample pairs $\left(\mathbf{S}_{i}, y_{i}\right)$ are streamed into the model one at a time, where $S_{i}$ is an RPM sample consisting of 8 context panels and 8 choice panels and $y_{i}$ is the associated label for one of the $K=8$ choice panels. The model mixes the current example with $r$ labeled samples, which are selected from the replay buffer based on a selection probability $p_{i}$ defined by:
\begin{equation}
    p_{i} = \frac{v_{i}}{\sum_{\mathbf{v}_{j} \in \mathcal{B}} v_{j}} \enspace ,
\end{equation}
where $v_{i}$ is the value associated with choosing sample $\mathbf{S}_{i}$ from buffer $\mathcal{B}$ for replay. Using an unlimited buffer, we study the seven selective replay policies described below.

\begin{figure*}[t]
\begin{center}
  \includegraphics[width=\textwidth]{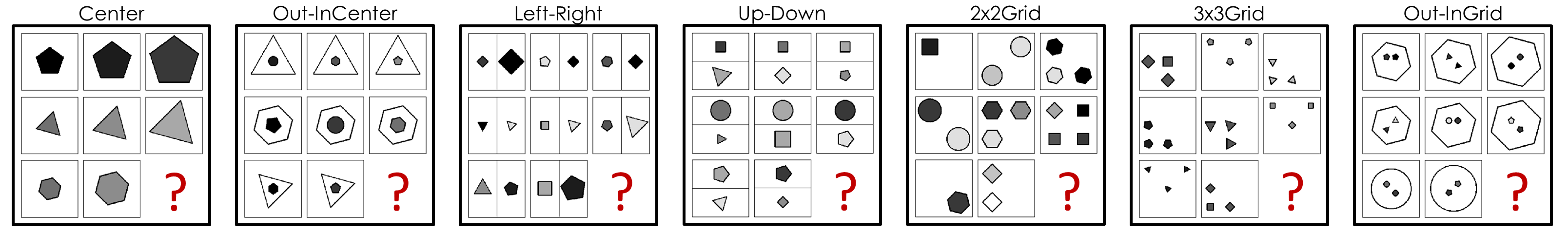}
\end{center}
  \caption{Example problems from each of the seven RAVEN tasks~\cite{zhang2019raven}.
}
\label{fig:raven-tasks}
\end{figure*}

\paragraph{Uniform Random:} Randomly select examples with uniform probability from the memory buffer. This is the simplest sampling approach and has demonstrated success for image classification~\cite{castro2018end,chaudhry2018riemannian,hayes2019remind,wu2019large}.
    
\paragraph{Minimum Logit Distance:}  Samples are scored according to their distance to a decision boundary~\cite{chaudhry2018riemannian}:
    \begin{equation}
        s_{i} = \sum_{j=1}^{K} \lvert\phi\left(\mathbf{S}_{i}\right)_{j}\mathbf{y}_{j}\rvert \enspace ,
    \end{equation}
    where $\phi\left(\mathbf{S}_{i}\right) \in \mathbb{R}^{K}$ is a vector of network scores for the current example and $\mathbf{y} \in \mathbb{R}^{K}$ is a one-hot encoding of the label, $y_{i}$. Since neural networks are more uncertain about samples close to decision boundaries, this method prioritizes replaying these difficult samples. 
    
\paragraph{Minimum Confidence:}  Samples are selected based on network confidence (i.e., softmax output):
    \begin{equation}
        s_{i} = \sum_{j=1}^{K} \textrm{softmax}\left(\phi\left(\mathbf{S}_{i}\right)\right)_{j}\mathbf{y}_{j} = P\left(C=y_{i}|\mathbf{S}_{i}\right) \enspace ,
    \end{equation}
    where $\textrm{softmax}\left(\cdot\right)$ returns the network's predicted probabilities and $\mathbf{y} \in C$ is a one-hot encoding of the label $y_{i}$. Intuitively, updating the model on samples that it is uncertain about could improve performance.
    
\paragraph{Minimum Margin:} Samples are selected by: 
    \begin{equation}
        s_{i} = P\left(C=y_{i}|\mathbf{S}_{i}\right) - \max_{y^{\prime}, y^{\prime} \neq y_{i}} P\left(C = y^{\prime}|\mathbf{S}_{i}\right)\enspace ,
    \end{equation}
    where the first term represents the probability of the classifier choosing the correct label and the second term represents the probability of the network choosing the most probable label from the remaining classes. Smaller margin values indicate more uncertainty.
    
\paragraph{Maximum Loss:} Scores are assigned based on cross-entropy loss:
    \begin{equation}
        s_{i} = -\sum_{j=1}^{K} \mathbf{y}_{j} \log P\left(j=y_{i}|\mathbf{S}_{i}\right)
        \enspace .
    \end{equation}
    Since networks seek to minimize classification loss, choosing to replay samples with the largest loss values should improve network performance.
    
\paragraph{Maximum Time Since Last Replay:} Samples are selected based on the last time they were seen by the network. Samples that have not been replayed in a long time could be forgotten and are prioritized for replay.
    
\paragraph{Minimum Replays:} Samples are selected based on the number of times they have been replayed to the network. Intuitively, samples with the fewest number of replays might not have been well-learned by the network and should be replayed. We initialize all replay counts to the number of base initialization epochs.

\vspace{\baselineskip}

To put all $s_{i}$ values into a similar range, we shift all values by the minimum value in the buffer such that the smallest value is 1, i.e., $s_{i} \leftarrow s_{i} + \left(1 - \min_{j}s_{j}\right)$. For the Random and Max Loss techniques, $v_{i}$ is equal to $s_{i}$. For all other techniques, we invert the $s_{i}$ values such that the most probable samples have the largest $v_{i}$ values, i.e., $v_{i} = \left({s_{i}+\epsilon} \right)^{-1}$, where $\epsilon=10^{-7}$ ensures the denominator is non-zero. 

After the $r$ samples are chosen, the network updates on this batch of $r+1$ samples for a single iteration and the associated $s_{i}$ values of the $r+1$ samples are subsequently updated. All $s_{i}$ values are appropriately initialized after the base initialization phase by pushing the base initialization data through the network. Further, during stream training, we only update the $s_{i}$ values for samples that were replayed to save on compute time.

For each selection policy, we evaluate two ways of choosing samples: unbalanced and balanced such that we oversample a task if it is underrepresented. While the unbalanced strategy replays samples strictly based on their selection probabilities, the balanced strategy ensures that replay samples are not prioritized from only a few classes.

\subsection{Implementation Details}

We use the hyperparameters and pre-processing steps for Rel-Base from \cite{spratley2020closer}. This includes resizing images to 80$\times$80, normalizing pixel values in $\left[0,1\right]$, and inverting images to increase signal. The hyperparameters are: Adam optimizer with learning rate=3e-4, $\beta_{1}$=0.9, $\beta_{2}$=0.999, $\epsilon$=1e-8, batch size=32, and epochs=50 per task for batch and base initialization models. For offline models, we run at least 50 epochs, at most 250 epochs, early stop if validation loss does not improve for 10 epochs, and choose the checkpoint with the highest validation accuracy. We use replay mini-batches of size 32 for our main selective replay experiments and compare additional batch sizes in Sec.~\ref{subsubsec:replay-sample-ablation}. For selective replay experiments, we allow models to use an unlimited buffer to focus on the selection methods directly. All models use a single output head, where task labels are unknown during test time. For regularization models, we grid searched for the regularization loss weight and found 1 and 10 to work best for Distillation and EWC, respectively. All timing experiments were run on the same machine with an NVIDIA TITAN RTX GPU, 48 GB of RAM, and an NVME SSD for consistency.

\section{Experimental Setup}
\label{sec:experimental-setup}

\subsection{Dataset \& Protocol}
\label{subsec:datasets}

We conduct experiments on the RAVEN dataset~\cite{zhang2019raven}, which has naturally defined tasks unlike the PGM dataset~\cite{barrett2018measuring}, making RAVEN more suitable for continual learning. RAVEN contains 1,120,000 images with 70,000 associated questions. These questions are distributed equally among seven unique figure configurations, depicted in Fig.~\ref{fig:raven-tasks}, where each configuration requires different reasoning capabilities. We use these configurations to define our continual learning tasks, i.e., each task consists of one unique configuration. Since the order in which tasks appear can impact performance, we evaluate models under several fixed task permutations and fix the sample order across models for all experiments. We run each experiment with the following three permutations of the task ordering and report the mean performance: \{Center, Out-InCenter, Left-Right, Up-Down, 2$\times$2Grid, 3$\times$3Grid, Out-InGrid\}, \{Up-Down, Center, Out-InCenter, Out-InGrid, 3$\times$3Grid, 2$\times$2Grid, Left-Right\}, and \{2$\times$2Grid, Left-Right, Out-InGrid, Up-Down, 3$\times$3Grid, Center, Out-InCenter\}. An example of our streaming protocol is in Fig.~\ref{fig:rpm-protocol}.

Statistical biases have been identified in RAVEN's answer set~\cite{hu2021stratified,spratley2020closer}, and \cite{spratley2020closer} suggested using models that process image frames independently to prevent bias exploitation. Since the Rel-Base model processes images independently, this bias is not a concern in our experimental results. 

\subsection{Metrics}
\label{subsec:metrics}

To compute our metrics, we define a matrix $R \in \mathbb{R}^{T \times T}$, where each entry $R_{i,j}$ denotes the continual learner's test accuracy on task $t_{j}$ after learning $t_{i}$ and there are $T$ total tasks. Following \cite{hayes2019memory,kemker2018forgetting}, we measure a continual learner's performance with respect to an offline baseline:
\begin{equation}
\label{eq:omega}
    \Omega = \frac{1}{T}\sum_{i=1}^{T}\frac{\gamma_{i}}{\gamma_{\mathrm{offline},i}} \enspace ,
\end{equation}
where $\gamma_{i}=\frac{1}{T}\sum_{j=1}^{T}R_{i,j}$ is the accuracy of the continual model at time $i$ and $\gamma_{\mathrm{offline},i}$ denotes the offline learner's accuracy at time $i$, computed using its associated $R_{\mathrm{offline}}$ matrix defined similarly to $R$. By normalizing the continual learner's performance to an offline learner, it is easier to compare results across task permutations. Higher values of $\Omega$ are better and an $\Omega$ of 1 indicates that the continual learner performed as well as the offline learner.

We also adopt three metrics from \cite{diaz2018forget} to evaluate average accuracy and backward/forward transfer. Using $R$, these metrics are defined as:
\begin{equation}
\label{eq:A}
A = \frac{2}{T\left(T+1\right)}{\sum_{i \geq j}^{T}{R_{i,j}}}
\end{equation} 
\begin{equation}
\label{eq:BWT}
BWT = \frac{2}{T\left(T-1\right)}{\sum_{i=2}^{T}{\sum_{j=1}^{i-1}{\left(R_{i,j}-R_{j,j}\right)}}} 
\end{equation}
\begin{equation}
\label{eq:FWT}
FWT = \frac{2}{T\left(T-1\right)}{\sum_{i<j}^{T}{R_{i,j}}} \enspace ,
\end{equation}
where $A$ is average model accuracy, $BWT$ is backward transfer, and $FWT$ is forward transfer. For $BWT$ and $FWT$, a larger value indicates that learning new tasks improved performance on previously seen tasks and unseen tasks, respectively. Note that $BWT$ can be negative, indicating that a model catastrophically forgot previous knowledge. We also report memory and compute requirements for each model since ideal learners require fewer resources.

\begin{table}[t]
\caption{Continual analogical reasoning performance on RAVEN. Each result is the average over three permutations. Additional memory requirements beyond the neural network (MB) and overall compute time (MIN) for each model are also reported. We report the best Partial Replay model (32 samples, Min Replays).
}
\label{tab:main-results}
\centering
\resizebox{\linewidth}{!}{
\begin{tabular}{lcccccc}
\toprule
\textsc{Model} & \textsc{$\Omega$} & \textsc{A} & \textsc{BWT} & \textsc{FWT} & \textsc{MIN} & \textsc{MB} \\ 
\midrule
\multicolumn{2}{l}{\textit{Stream Learners}}\\
Fine-Tune & 0.256 & 0.121 & -0.238 & 0.091 & 8 & 0 \\
Partial Replay & 0.924 & 0.811 & 0.006 & 0.229 & 65 & 4301 \\
\midrule
\multicolumn{2}{l}{\textit{Batch Learners}}\\
Fine-Tune & 0.581 & 0.417 & -0.456 & 0.183 & 77 & 0 \\
Distillation & 0.513 & 0.357 & -0.288 & 0.167 & 94 & 5 \\
EWC & 0.615 & 0.459 & -0.347 & 0.178 & 85 & 5 \\
Cumul. Replay & \textbf{0.990} & \textbf{0.893} & \textbf{0.028} & \textbf{0.232} & 325 & 4301 \\
\bottomrule
\end{tabular}
}
\end{table}

\section{Results}
\label{sec:results}

\subsection{Main Results}
\label{subsec:baseline-results}

\begin{figure}[t]
\begin{center}
  \includegraphics[width=0.8\linewidth]{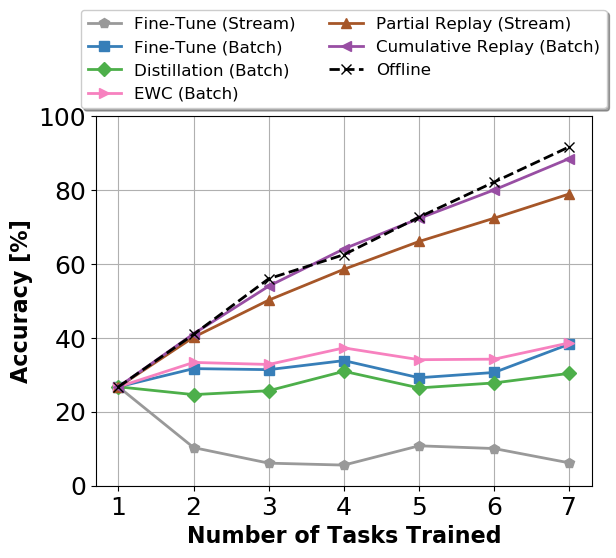}
\end{center}
  \caption{Learning curves for each baseline model.
}
\label{fig:learn-curve}
\end{figure}

Our main results are in Table~\ref{tab:main-results} and the associated learning curves are in Fig.~\ref{fig:learn-curve}. Chance performance is \sfrac{1}{8} (12.5\%), which is equal to an $\Omega$ of 0.237. The final accuracy that we use to normalize $\Omega$ is 91.7\%, as reported in \cite{spratley2020closer}. We include the top performing Partial Replay model (using the unbalanced Min Replays selection strategy) for comparison.

In terms of performance measures, Cumulative Replay consistently performed the best, with the Partial Replay model performing second best and requiring less time. Since the Partial Replay model uses an unlimited replay buffer, its memory usage is tied for worst with the Cumulative Replay learner. In the future, different buffer management strategies could be evaluated and paired with the best selective replay strategies to improve performance and reduce memory resources further. While the streaming Fine-Tune method required the fewest computational resources, it had the worst overall performance in terms of $\Omega$, $A$, and $FWT$. This is unsurprising as Fine-Tune (Stream) does not have any mechanisms to mitigate catastrophic forgetting and sees each training example only once. The batch variant of Fine-Tune performed much better than the streaming variant, but had the worst overall $BWT$. We hypothesize this is because Fine-Tune (Batch) does not have any mechanisms to mitigate forgetting and batch training on new tasks causes overfitting, leading to worse $BWT$.

EWC outperformed both Fine-Tune variants in terms of $\Omega$, $A$, and $BWT$, but had slightly worse $FWT$ than Fine-Tune (Batch), likely due to its regularization loss. Distillation performed worse than Fine-Tune (Batch). While Distillation has demonstrated success in standard supervised classification scenarios, the distillation penalty does not prove useful for continual learning on RPMs. This could be because the output space for RPMs consists of a multiple choice problem of selecting one of eight images, which doesn't have as much semantic meaning as explicit object categories in standard classification settings. Note that both regularization models and Fine-Tune models suffered from forgetting, as indicated by their negative $BWT$ scores.

\subsection{Selective Replay Results}
\label{subsec:selective-replay-results}

We endeavor to study the impact of selective replay policies on Partial Replay performance using the strategies outlined in Sec.~\ref{sec:sample-selection}. Performance for each of these methods in the unbalanced and balanced settings are in Table~\ref{tab:raven-select}. Overall, the top performing method was Min Replays and the worst performing method was Random. We ran multiple comparisons tests against both of these selection methods (Random and Min Replays) in the unbalanced and balanced settings to determine if the other selection methods were statistically significantly different. To perform the tests, we first sampled 300 non-overlapping subsets of test instances randomly and computed the average final accuracy of the subsets across the three runs. We then ran a paired Welch's t-test (unequal variances) on the sets of subset accuracies. Unless otherwise noted, we corrected for multiple hypothesis testing using Holm-Bonferroni correction and used a significance level of $0.01$.

In the unbalanced case, all selective replay methods had statistically significant performance differences compared to the Random selection policy, and we found that all selective replay strategies had statistically significant performance differences compared to the Min Replays policy (for all comparisons, $P < 0.001)$. In the balanced case, only the Min Margin and Min Replays methods had statistically significant performance differences compared to the Random baseline (for both, $P < 0.001)$, and only Random was statistically significant from Min Replays ($P < 0.001)$.

We also performed significance tests of each unbalanced selection method with its associated balanced counterpart and found that only the unbalanced Max Loss and Min Replays strategies were statistically significant from their balanced variants (without Holm-Bonferroni correction). We also ran a paired t-test of an average of the final results from all selection methods in the unbalanced versus balanced settings and failed to reject the null hypothesis of equal means without Holm-Bonferroni correction ($P = 0.08$). Thus, on average there is no statistical significance between the unbalanced and balanced sampling strategies.

\begin{figure}[t]
    \centering
        \includegraphics[width=\linewidth]{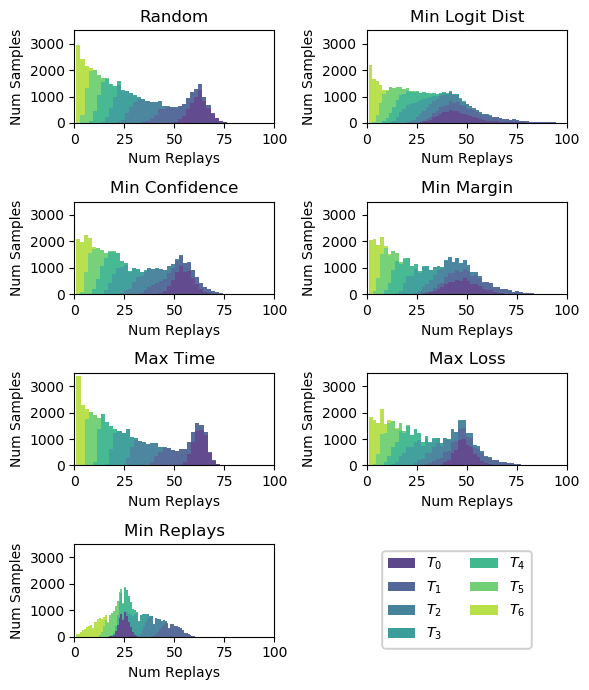}
    \caption{Histograms for each unbalanced selective replay method showing the number of samples with their associated replay counts after streaming learning. Each color denotes samples from the $i$-th task ($T_{i}$). Each plot is the average over three order permutations.
    }
    \label{fig:selective-replay-hist}
    \vspace{-0.15mm}
\end{figure}

Fig.~\ref{fig:selective-replay-hist} shows histograms of the number of training samples with their associated number of replays after the completion of stream training for each unbalanced selection method. Qualitatively, the histograms for the Random and Max Time strategies, which performed the worst, look similar. Both histograms have the most samples with the fewest replays across all histograms and also have the most replays of the first task. These results suggest the poor performance of these methods was due to overplaying a small set of examples, while underplaying many other examples. Visually, the Min Confidence, Min Margin, Min Logit Dist, and Max Loss replay distributions look similar and performed similarly. The most unique distribution is from the top-performing Min Replays strategy, which has the fewest samples with the fewest replays and a more uniform replay count across all tasks compared to the other histograms.

\begin{table}[t]
\caption{Comparison of selective replay strategies with unbalanced (unbal) and balanced (bal) sampling, averaged over three runs. Final accuracies (FA) used for significance testing are also reported.
}
\label{tab:raven-select}
\centering
\resizebox{\linewidth}{!}{
\begin{tabular}{lcccccc}
\toprule
& \multicolumn{3}{c}{\textsc{Unbal}} & \multicolumn{3}{c}{\textsc{Bal}} \\ 
\cmidrule(r){2-4} \cmidrule(r){5-7} 
\textsc{Method} & \textsc{$\Omega$} & \textsc{A} & \textsc{FA} & \textsc{$\Omega$} & \textsc{A} & \textsc{FA} \\
\midrule
Random & 0.882 & 0.769 & 0.752 & 0.897 & 0.785 & 0.758\\
Min Logit Dist & 0.905 & 0.793 & 0.772 & 0.895 & 0.785 & 0.764 \\
Min Confidence & 0.895 & 0.783 & 0.764 & 0.902 & 0.790 & 0.767 \\
Min Margin & 0.906 & 0.795 & 0.773 & 0.900 & 0.790 & \textbf{0.773} \\
Max Time & 0.887 & 0.774 & 0.764 & 0.897 & 0.787 & 0.762 \\
Max Loss & 0.909 & 0.800 & 0.776 & 0.900 & 0.789 & 0.763 \\
Min Replays & \textbf{0.924} & \textbf{0.811} & \textbf{0.790} & \textbf{0.907} & \textbf{0.795} & 0.771 \\
\bottomrule
\end{tabular}
}
\end{table}

\subsubsection{Influence of Number of Replay Samples}
\label{subsubsec:replay-sample-ablation}

\begin{figure}[t]
\begin{center}
  \includegraphics[width=0.8\linewidth]{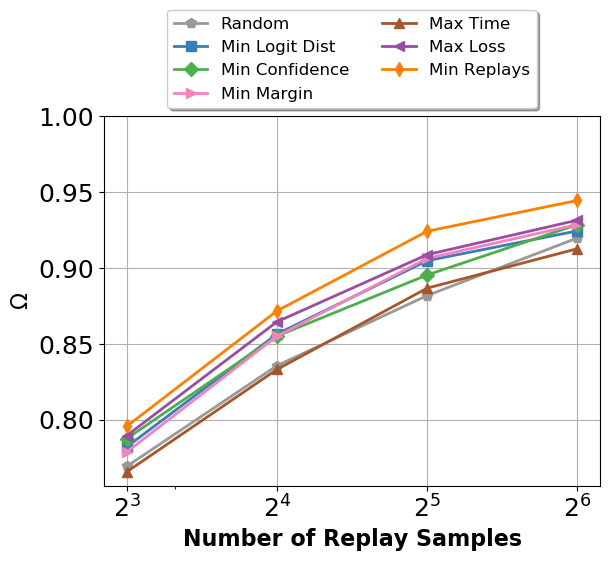}
\end{center}
  \caption{$\Omega$ performance as a function of replay samples for each unbalanced selective replay method averaged over three runs.
  }
\label{fig:num-replay-samples}
\end{figure}

Our main Partial Replay experiments used replay mini-batches of size 32. However, we were also interested in how each selection method's performance changed as a function of mini-batch size. Overall $\Omega$ results for each method using replay batches of size 8, 16, 32, and 64 are shown in Fig.~\ref{fig:num-replay-samples}. All curves are monotonically increasing with the Min Replays and Max Loss strategies yielding the top two performances across all batch sizes. Similarly, Random and Max Time produced the worst results across all batch sizes. Although there was a slight average performance increase (2.6\% $\Omega$) across all methods from a batch size of 32 to 64, running experiments with 64 samples required 1.9$\times$ more compute time, making it less ideal for streaming learning.

\section{Discussion}
\label{sec:discussion}

Our main results indicate that replay-based learners perform the best for continually solving RPM puzzles. Although Cumulative Replay completely mitigates forgetting, it is computationally intensive and not ideal for streaming learning. To overcome this compute bottleneck, researchers often use Partial Replay to replay only a subset of the dataset at each time-step. Our experimental results indicate that Partial Replay performs well and its performance can be further improved by strategically selecting samples based on some criteria. While all sample selection methods we tested performed better than uniform random selection, replaying samples based on the Min Replays and Max Loss strategies yielded the best overall results. In the future, it would be interesting to explore sample selection methods that optimize for samples directly~\cite{liu2020mnemonics}, optimize for selection directly~\cite{aljundi2019online}, or train a teacher network to choose samples for the learner~\cite{fan2018learning}. In an offline setting, \cite{zheng2019abstract} paired a reinforcement learning policy with a teacher to intelligently compose a training curriculum for an RPM-based student, which improved performance over standard mini-batch training. This is promising evidence to explore additional selective replay strategies for continual RPM learning. Additionally, future work should include testing additional architectures designed for RPMs in the continual setting, which was beyond the scope of this study. This would inform future continual learning model designs.

Beyond RPMs, it would also be interesting to explore additional problems for continual analogical reasoning, including those introduced in \cite{hill2018learning}. Moreover, general abstract reasoning requires additional skills such as numerical reasoning, inductive reasoning, logical reasoning, etc. Future studies could explore continual learning in the context of these other reasoning skills using baselines introduced in \cite{hernandez2016computer,saxton2018analysing}. While most of these tasks require only reasoning, agents operating in the real-world should also be capable of processing additional inputs such as natural language questions, or identifying and mitigating biases to abstract general knowledge. Some problems requiring these additional components include visual question answering~\cite{antol2015vqa,kafle2016review,malinowski2014multi}, referring expression recognition~\cite{kazemzadeh2014referitgame,rohrbach2016grounding}, visual query detection~\cite{acharya2019vqd}, and image captioning~\cite{bernardi2016automatic}. While models have been developed for continual visual question answering~\cite{greco-2019-psycholinguistics,hayes2019remind}, the abstraction capabilities of these models have not been evaluated directly. More studies should be conducted to evaluate models on additional reasoning tasks.

\section{Conclusion}
\label{sec:conclusion}

While humans continually acquire new information and strengthen their reasoning capabilities over their lifetimes, deep neural networks struggle with these problems. In this paper, we introduced protocols, baseline methods, and metrics for evaluating networks on continual analogical reasoning tasks using the RPM-based RAVEN dataset. We found that replay methods had the best global performance and backward/forward knowledge transfer. We further studied several replay selection policies and found statistically significant performance improvements by all methods over a uniform random policy. Designing and testing more sophisticated architectures and continual learning strategies for RPMs remains an area of future work.

\paragraph{Acknowledgements.}
This work was supported in part by the DARPA/SRI Lifelong Learning Machines program [HR0011-18-C-0051], AFOSR grant [FA9550-18-1-0121], and NSF award \#1909696. The views and conclusions contained herein are those of the authors and should not be interpreted as representing the official policies or endorsements of any sponsor. We thank Robik Shrestha and Jhair Gallardo for their comments and useful discussions.

{\small
\bibliographystyle{ieee_fullname}
\bibliography{egbib}
}

\end{document}